\documentclass[twoside]{article}

\usepackage[accepted]{aistats2024}
\usepackage[round]{natbib}

\usepackage[
    pagebackref=true,
    linkcolor=red!80!black,
    citecolor=blue!70!black,
    urlcolor=blue!70!black,
    colorlinks,
    breaklinks
]{hyperref}

\usepackage{amsmath}
\usepackage{amsthm}
\usepackage{amssymb}
\usepackage{mathrsfs}
\usepackage{booktabs}
\usepackage{algorithm}
\usepackage{algorithmic}
\usepackage{graphicx}
\usepackage{subfigure}
\usepackage{bbding}
\usepackage{multirow}
\usepackage{xcolor} 
\usepackage{mathtools,color}
\newtheorem{assumption}{Assumption}
\newtheorem{definition}{Definition}
\newtheorem{proposition}{Proposition}
\newtheorem{theorem}{Theorem}

\newtheorem{problem}{Problem}

\newtheorem*{remark}{Remark}

\begin{document}

%

%
\runningauthor{Jiashuo Liu, Jiayun Wu, Jie Peng, Xiaoyu Wu, Yang Zheng, Bo Li, Peng Cui}

\twocolumn[

\aistatstitle{Enhancing Distributional Stability among Sub-populations}

\aistatsauthor{ Jiashuo Liu$^{1,*}$, Jiayun Wu$^1$, Jie Peng$^1$, Xiaoyu Wu$^2$, Yang Zheng$^2$, Bo Li$^3$, Peng Cui$^{1,\dagger}$}

\aistatsaddress{ 
$^1$Department of Computer Science and Technology, Tsinghua University\\ 
	$^2$RAMS Lab, Huawei Technologies Co Ltd\\
	$^3$School of Economics and Management, Tsinghua University} 
	\vspace{-2em}
\aistatsaddress{$^*$\texttt{liujiashuo77@gmail.com, $^\dagger$cuip@tsinghua.edu.cn}}]

\begin{abstract}
  Enhancing the stability of machine learning algorithms under distributional shifts is at the heart of the Out-of-Distribution (OOD) Generalization problem. Derived from causal learning, recent works of invariant learning pursue strict invariance with multiple training environments. Although intuitively reasonable, strong assumptions on the availability and quality of environments are made to learn the strict invariance property. In this work, we come up with the ``distributional stability" notion to mitigate such limitations. It quantifies the stability of prediction mechanisms among sub-populations down to a prescribed scale. Based on this, we propose the learnability assumption and derive the generalization error bound under distribution shifts.
  Inspired by theoretical analyses, we propose our novel stable risk minimization (SRM) algorithm to enhance the model's stability w.r.t. shifts in prediction mechanisms ($Y|X$-shifts). Experimental results are consistent with our intuition and validate the effectiveness of our algorithm.
  The code can be found at \url{https://github.com/LJSthu/SRM}.
\end{abstract}

\section{INTRODUCTION}
Traditional machine learning algorithms with empirical risk minimization (ERM) are vulnerable when exposed to data drawn out of the training distribution.
In order to mitigate the failures in the out-of-distribution (OOD) generalization, invariant learning methods~\citep{IRM, IRMgames, IB, ICP} are proposed to learn prediction mechanisms that are strictly invariant across given multiple environments.
Such strict invariance property enables models to generalize under distributional shifts \citep{ICP, causalTransfer, IRM, MIP}.

To fulfill the promise of invariant learning, \emph{environment labels} have to be provided to achieve strict invariance.
Moreover, the concept of strict invariance even assumes the access of \emph{all possible environments}.
However, such requirement is unrealistic in real-world applications where modern datasets are often constructed by amalgamating data from various sources, thus significantly limiting the applicability of invariant learning techniques.
Recent efforts, such as EIIL~\citep{EIIL} and HRM~\citep{hrm,liu2021integrated}, have focused on generating pseudo environment labels to facilitate invariant learning. 
Nonetheless, the characteristics of these pseudo environments, the extent of invariance they enable, and even the validity of the problem framework itself, remain unclear and inadequately justified.

To address these limitations, our research shifts focus towards developing models that generalize out-of-distribution within contexts of latent heterogeneity, where the training data is gathered from multiple sources, but lacking explicit source labels. 
In this setting, the training data exhibits sub-population structures, with probably distinct prediction mechanisms varying across sub-populations.
To tackle this problem, we introduce an approach that extends strict invariance to the concept of ``distributional stability". 
This metric assesses the consistency of prediction mechanisms across sub-populations. 
Unlike the binary nature of strict invariance, which is either yes or no, distributional stability provides a continuous measure that quantifies the degree of predictive mechanism stability across varying contexts. 
This nuanced approach allows for a more refined assessment of model robustness in handling distribution shifts.

In Section~\ref{sec:distributional-invariance}, we formally define the distributional stability, and introduce its properties as well as relationships with strict invariance.
And we also demonstrate its relationship with the distributional robustness from the distributionally robust optimization literature \citep{fDRO, duchi2019distributionally}.
Then in Section \ref{sec:theorem},
we characterize the \emph{learnability} of the problem to rationalize the problem setting itself and clarify what kind of target distributions could be generalized to.
Then we derive the \emph{OOD generalization error bound} for this problem based on the distributional stability.
Inspired by the theoretical results, we find that models with strong distributional stability could generalize well with respect to shifts on prediction mechanisms ($Y|X$-shifts).
Thus, we propose an empirical algorithm named \emph{Stable Risk Minimization} (SRM) in Section \ref{sec:methods}, and experimental results on both simulation and real-world data validate the effectiveness of our method.

\textbf{Notations}\quad Throughout this paper, we let $X\in\mathcal{X}$ denote the covariates, $Y\in\mathcal{Y}$ denote the target.
$f_\theta(\cdot):\mathcal{X}\rightarrow\mathcal{Y}$ is the predictor parameterized by $\theta\in\Theta$. 
$\mathcal{E}$ is the random variable taking values in all possible environments.
The random variable of data points is denoted by $Z=(X,Y)\in\mathcal{Z}$.
$\mathbb P^e(Z)$ abbreviated with $\mathbb P^e$ denotes the joint distribution in environment $e$, and for environments $e_1,e_2\in \text{supp}(\mathcal{E})$, the data distribution can be quite different.
$\mathbb  P_{\text{train}}(Z)$ and $\mathbb P_{\text{test}}(Z)$ abbreviated with $\mathbb P_{\text{tr}}$ and $\mathbb P_{\text{te}}$ respectively represent the joint training distribution and test distribution.
Denote the feature extractor $\Phi_\theta(X)$ parameterized by $\theta$, and the predicting function $\hat{Y}=h_\eta(\Phi_\theta(X))$ parameterized by $\eta$ (not restricted to linear $h(\cdot)$), which gives the whole prediction model $f_{\eta,\theta}(X)=h_\eta(\Phi_\theta(X))$.
For simplicity, we omit the subscripts $\theta,\eta$ without causing misunderstanding.
Denote the sample size by $n$ and the vector of sample weights $\textbf{w}\in\mathbb{R}_+^n=[w_1, \dots, w_n]^T$ with $\textbf{w}\geq 0$ and $\textbf{w}^T\textbf{1}=1$.

\section{DISTRIBUTIONAL STABILITY}
\label{sec:distributional-invariance}
In this section, we first introduce the \emph{strict invariance} property as well as its limitations.
Then we propose the \emph{distributional stability} property, a relaxed alternative under latent heterogeneity.

\subsection{Strict invariance}

Inspired by causal inference literature, strict invariance~\citep{IRM, IRMgames, MIP,EIIL, hrm, liu2021integrated} requires  that the prediction mechanism $Y|X$ remains the same among environments, which has two typical formulations.

\begin{definition}[Strict Invariance]
\label{def:invarianceMIP}
Denote the random variable taking values of all possible environments as $\mathcal{E}$.
A representation $\Phi$ is strictly invariant if condition 1 or condition 2 holds:\\
Condition 1~\citep{IRM, IRMgames, EIIL}:  for any $e_1,e_2\in\text{supp}(\mathcal{E})$,
\begin{equation}
\label{equ:condition1}
	\mathbb{E}[Y|\Phi,\mathcal{E}=e_1]=\mathbb{E}[Y|\Phi,\mathcal{E}=e_2].
\end{equation}
Condition 2~\citep{MIP, hrm, liu2021integrated}: for any $e_1,e_2\in\text{supp}(\mathcal{E})$,
\begin{equation}
\label{equ:condition2}
	\mathbb P(Y|\Phi,\mathcal E=e_1)=\mathbb P(Y|\Phi, \mathcal E=e_2).
\end{equation}
\end{definition}
Invariant learning methods use strict invariance as a constraint during the model learning procedure.
\cite{IRM} prove that a linear model only uses invariant features under condition \eqref{equ:condition1}, and \cite{MIP} prove the resultant model under condition \eqref{equ:condition2} is optimal for OOD generalization.
Despite the promising theoretical results, one major concern in defining the strict invariance as Definition \ref{def:invarianceMIP} is the access to \emph{all possible environments} $\mathcal E$.

The strict invariance requires all possible environments to examine whether the prediction mechanism $Y|\Phi$ stays invariant.
However, in most of the real-world applications, it is impossible to acquire all possible environments, which renders the goal of strict invariance unrealistic to reach in practice.
As a result, the learned invariance only holds for the finite training environments, but \emph{whether} it is violated in other agnostic environments and \emph{how much} it is violated remain entirely unknown for machine learning engineers and system users, which brings huge risks in high-stakes applications.

\subsection{A relaxed alternative}
To mitigate the limitations above, we relax the requirements for multiple environments and instead consider an elaborated setting where the observed data are heterogeneous.
More precisely, following \citet{duchi2019distributionally}, we assume that 
\begin{equation*}
	X,Y\sim \mathbb P_{\text{tr}} \coloneqq \alpha \mathbb Q_0 + (1-\alpha)\mathbb Q_1
\end{equation*}	
where the proportion $\alpha\in (0,1)$ and $\mathbb Q_0,\mathbb Q_1$ denote the sub-populations in $\mathbb P_{\text{tr}}$.
Since the sub-population distributions are not pre-defined, it is termed latent heterogeneity.
To measure the stability of a machine learning model under potential distributional shifts, inspired by strict invariance among given environments (i.e. explicit heterogeneity), we could examine whether the predicting mechanism holds among all potential sub-populations within $\mathbb P_{\text{tr}}$.
First, we define the sub-population set for a distribution in Definition \ref{def:set}.

\begin{definition}[Sub-population set]
\label{def:set}
Given distribution $\mathbb P(Z)$, for $\alpha_0\in(0,1/2)$ as a lower bound  on the sub-population proportion $\alpha$, the set of sub-populations of distribution $\mathbb P$ is 
	\begin{equation*}
	\begin{aligned}
		\mathcal{P}_{\alpha_0}(\mathbb P)\coloneqq \{\mathbb Q_0: \mathbb P=\alpha \mathbb Q_0+(1-\alpha)\mathbb Q_1,\text{ for some }\\
		\alpha\in[\alpha_0,1) \text{ and distribution }\mathbb Q_1 \text{ on }\mathcal{Z} \}.
	\end{aligned}
	\end{equation*}	
\end{definition}

\begin{remark}	

Intuitively, $\mathcal P_{\alpha_0}(\mathbb P)$ contains all sub-populations of $\mathbb P$ with proportion $\alpha\geq \alpha_0$. $\alpha_0$ controls the size of the minimal sub-populations considered, i.e. smaller $\alpha_0$ corresponds with smaller sub-populations but larger size of the set ($|\mathcal{P}_{\alpha_0}|$).
\end{remark}

Based on this, we introduce a nuanced variant of strict invariance, named $\alpha_0$-distributional stability. 
This concept serves to quantify the level of stability in the face of shifts among sub-populations. 
It represents a more flexible approach that captures the degree to which a model's predictions remain consistent across varying sub-populations, offering a refined metric for evaluating the robustness of models in heterogeneous data environments.

\begin{definition}[$\alpha_0$-distributional stability]
\label{def:distributionalinvariance}
Given data distribution $\mathbb P(Z)$, for $\alpha_0\in(0,1/2)$, the $\alpha_0$-distributional stability of the prediction mechanism $Y|X$ is defined as 
\begin{equation}
\label{equ:def}
\text{DS}_{\alpha_0}(Y|X;\mathbb P)\coloneqq	\sup_{\mathbb Q\in\mathcal{P}_{\alpha_0}(\mathbb P)}\rho_{\text{KL}}(\mathbb Q(Y|X), \mathbb P(Y|X))
\end{equation}
where $\rho_{\text{KL}}(\cdot,\cdot)$ denotes the KL-divergence between two distributions.
\end{definition}

\begin{remark}
Intuitively, $\alpha_0$-distributional stability measures the \emph{maximal variation} of the prediction mechanism  ($Y|X$) among sub-populations within $\mathbb P$ in terms of KL-divergence.
It picks the worst sub-population $\mathbb Q^\star$ in the set $\mathcal P_{\alpha_0}(\mathbb P)$ and calculates the KL-divergence between $\mathbb Q^\star(Y|X)$ and $\mathbb P(Y|X)$.
The smaller the $\text{DS}_{\alpha_0}$ is, the more stable the prediction mechanism $Y|X$ is, since one can hardly find a sub-population that violates $\mathbb P(Y|X)$.
\end{remark}

Then we demonstrate some properties of the proposed $\alpha_0$-distributional stability.

\begin{proposition}[Properties of DS$_{\alpha_0}(\mathbb P)$]
	For observed data distribution $\mathbb P(Z)$ and $\alpha_0\in (0,1/2)$, we have
	
	1. Nonnegativity: DS$_{\alpha_0}(Y|X;\mathbb P)\geq 0$;
	
	2. Monotonicity: if $\alpha_1 \geq \alpha_2$, we have DS$_{\alpha_1}(Y|X;\mathbb P) 
	\leq \text{DS}_{\alpha_2}(Y|X;\mathbb P)$ 
\end{proposition}

\begin{remark}
The smaller $\alpha_0$ is, the larger distribution set $\mathcal{P}_{\alpha_0}(\mathbb P)$ is, and the larger the stability criterion is, since the mechanism $Y|X$ is examined under more fine-grained sub-populations.	
\end{remark}

\begin{proposition}[Relationship with strict invariance]
	Here we demonstrate the connections and differences between $\alpha_0$-distributional stability and strict invariance:
	
	1. Connection with condition \eqref{equ:condition1}: replace $\rho_{\text{KL}}(\cdot,\cdot)$ with
	$\mathbb E[\|\mathbb{E}_{\mathbb Q}[Y|X]-\mathbb{E}_{\mathbb P}[Y|X]\|^2]$, and replace the sub-population set $\mathcal{P}_{\alpha_0}(\mathbb P)$ with $\mathcal{E}$,  then we have: $\text{DS}_{\alpha_0}(Y|X;\mathbb P)=0$ is equivalent to condition \eqref{equ:condition1}.
	
	2. Connection with condition \eqref{equ:condition2}: replace the sub-population set $\mathcal{P}_{\alpha_0}(\mathbb P)$ with $\mathcal{E}$, then we have:
	$\text{DS}_{\alpha_0}(Y|X;\mathbb P)=0$ is equivalent to condition \eqref{equ:condition2}.
\end{proposition}

\begin{remark}[Connection with distributional robustness]
\label{remark}
	Although both terms involve the sub-population set, distributional stability and distributional robustness are inherently different from each other.
	Distributional robustness~\citep{fDRO, Certifying, duchi2019distributionally} refers to the \emph{worst-case performance} inside the pre-defined uncertainty set $\mathcal{P}$, while distributional stability measures the \emph{maximal variation} of the prediction mechanism $Y|X$.
	Therefore, distributional robustness reflects the performance at a single point (i.e. the worst-case distribution), but distributional stability measures the variation of the prediction mechanisms (i.e. contrast between two distributions).
	Such difference leads to a huge discrepancy in the guarantees of the OOD generalization performances.
	DRO methods to obtain distributional robustness could \emph{only} ensure the performance within the distribution set $\mathcal P$, while methods to pursue distributional stability could \emph{generalize to agnostic testing distributions} under the \emph{learnability assumption}, which will be discussed in detail in Section \ref{sec:theorem}.
\end{remark}

\section{THEORETICAL ANALYSIS}
\label{sec:theorem}
Based on \emph{distributional stability}, we formally define the OOD generalization problem under latent heterogeneity.
Then we provide theoretical analysis of this problem, including the learnability assumption and the generalization error bound.

\begin{problem}[Setup]
\label{problem}
	Given data $Z\sim \mathbb P_{\text{tr}}(Z)$ collected from \textbf{multiple agnostic sources}, the goal is to learn models with good generalization performances on data from agnostic target distribution $\mathbb P_{\text{te}}(Z)$.
\end{problem}

For traditional machine learning problems, the analysis of the learnability is based on the $i.i.d.$ assumption.
However, in Problem \ref{problem}, the target distribution is agnostic and could significantly differ from the training one.
Therefore, without any further assumptions, even the learnability itself can \emph{hardly} hold in general.
Given this, we characterize the learnability assumption of Problem \ref{problem}, which makes assumptions on the target distribution.
Following \cite{Theoretical}, we define the expansion function as follows:
\begin{definition}[Expansion Function]
	A function $s:\mathbb{R}^+\cup \{0\}\rightarrow\mathbb{R}^+\cup\{0, +\infty\}$ is an expansion function, \textit{iff} the following properties hold: (1) $s(\cdot)$ is monotonically increasing and $s(x)\geq x, \forall x$; (2) $\lim\limits_{x\rightarrow 0^+}s(x)=s(0)=0$. 
\end{definition}

Besides, for training distribution $\mathbb P_{\text{tr}}(Z)$ and target distribution $\mathbb P_{\text{te}}(Z)$, we define the \emph{out-of-distribution stability} of $Y|X$ as:
	\begin{equation*}
		\text{ODS}(Y|X;\mathbb P_{\text{tr}},\mathbb P_{\text{te}})\coloneqq \rho_{\text{KL}}(\mathbb P_{\text{te}}(Y|X),\mathbb P_{\text{tr}}(Y|X)),
	\end{equation*}	
which measures the stability of the prediction mechanism between $\mathbb P_{\text{tr}}$ and $\mathbb P_{\text{te}}$.
Note that here $\mathbb P_{\text{te}}$ denotes the target distribution, which may not be included in the pre-defined sub-population set.

Then we formally provide the learnability assumption of Problem \ref{problem}.

\begin{assumption}[Learnability of Problem 1]
\label{def:invariance}
	Problem \ref{problem} from $\mathbb P_{\text{tr}}$ to $\mathbb P_{\text{te}}$ is $(\alpha_0,s)$-learnable if there exists an expansion function $s(\cdot)$ such that $\text{ODS}(Y|X;\mathbb P_{\text{tr}},\mathbb P_{\text{te}})\leq s(\text{DS}_{\alpha_0}(Y|X;\mathbb P_{\text{tr}}))$.
\end{assumption}

Note that here $X$ could be replaced by some representations $\Phi(X)$.
Here we make some remarks.

\begin{remark}
	\textbf{(1)} Assumption \ref{def:invariance} assumes that the $\alpha_0$-distributional stability measure of the training distribution should \emph{approximately hold} in testing, that is, its variation on the target distribution is upper bounded by the expansion function. 
	Intuitively, it requires that the conditional distribution $\mathbb P_{\text{te}}(Y|X)$ cannot arbitrarily change.
	If $\mathbb P_{\text{te}}(Y|X)$ could arbitrarily change, the problem is unlearnable, since the prediction mechanism learned in training may not hold in testing.\\
\textbf{(2)} The steepness of the expansion function reflects the \emph{difficulty} of Problem \ref{problem}, since the steeper the expansion function is, the less likely the learned distributional stability will hold in testing.
	As shown in Theorem \ref{theorem:bound}, the expansion function influences the generalization error bound.
\end{remark}

We then derive the OOD generalization bound for Problem \ref{problem}.

\begin{theorem}[Generalization Bound]
	\label{theorem:bound}
Under Assumption \ref{def:invariance}, assume that $\ell(\cdot,\cdot)$ is upper bounded, the conditional generalization error gap could be bounded by the distributional stability as:
	\begin{equation}
	\label{equ:bound1}
	\begin{aligned}
		\mathbb{E}_{\mathbb P_{\text{te}}}\bigg[\Big\|\mathbb{E}_{\mathbb P_{\text{te}}}[\ell(X,&Y)|X]-\mathbb{E}_{\mathbb P_{\text{tr}}}[\ell(X, Y)|X]\Big\|\bigg]\\
		&\leq \mathcal O(\sqrt{1-e^{-s(\text{DS}_{\alpha_0}(Y|X;\mathbb P_{\text{tr}}))}}),
	\end{aligned}
	\end{equation}	
	where $\ell(\cdot,\cdot)$ denotes the loss function.
\end{theorem}

In Theorem \ref{theorem:bound}, we calculate the \emph{conditional} error gap bound, which excludes covariate shifts by aligning the covariate distribution with $\mathbb P_{\text{te}}(X)$.
From Equation \eqref{equ:bound1}, we can see that controlling the distributional stability $\text{DS}_{\alpha_0}(Y|X;\mathbb P_{\text{tr}})$ could decrease the generalization error gap between training and testing.
The theoretical results motivate our Stable Risk Minimization (SRM) algorithm in Section \ref{sec:methods}.

\section{METHOD}
\label{sec:methods}
To enhance the distributional stability, inspired by Theorem \ref{theorem:bound}, we propose our \textbf{Stable Risk Minimization} (SRM) algorithm based on the newly-proposed distributional stability.
We first introduce the overall objective function, and then derive an approximated optimization method for classification and regression.\\

\textbf{Objective function.}

To learn models with good distributional stability, we introduce the stability constraints to the general risk minimization and propose our stable risk minimization framework as:
\begin{equation}
\label{equ:objective}
\begin{aligned}
	\theta^*,\eta^* &= \arg\min_{\theta,\eta} \mathbb{E}_{X,Y\sim \mathbb P_{\text{tr}}}\left[\ell(h_{\eta}(\Phi_{\theta}(X)),Y)\right]\\
	&\text{s.t.\quad\quad DS}_{\alpha_0}(Y|\Phi_{\theta^*}(X);\mathbb P_{\text{tr}})\leq \delta
\end{aligned}		
\end{equation}
where $\alpha_0$ is the pre-defined lower-bound on the sub-population proportion, and $\delta\geq 0$ is the threshold of distributional stability of the prediction mechanism $Y|\Phi_{\theta^*}(X)$.
The constraint could help to learn representation $\Phi_{\theta^*}(X)$ that is stable among sub-populations within $\mathbb P_{\text{tr}}$.

Following the approximation techniques typically adopted in robust learning \citep{IRM, Certifying}, we give up the requirement of a prescribed constraint $\delta$ of distributional stability, and instead focus on the Lagrangian penalty problem, which also corresponds with our theoretical results in Theorem \ref{theorem:bound}.
\begin{equation}
	\label{equ:lagrangian}
	\min_{\theta,\eta} \mathbb{E}_{\mathbb P_{\text{tr}}}[\ell(h_{\eta}(\Phi_{\theta}(X)),Y)] + \lambda\cdot\text{DS}_{\alpha_0}(Y|\Phi_{\theta}(X);\mathbb P_{\text{tr}})
\end{equation}
The \emph{key challenge} lies in the calculation of the distributional stability constraint $\text{DS}_{\alpha_0}(Y|\Phi_\theta(X);\mathbb P_{\text{tr}})$.
Recall that it relies on the worst sub-population $\mathbb Q^\star$ in Equation \eqref{equ:def}.
Therefore, the optimization involves a \emph{two-player} game, where a \emph{variation exploiter} keeps picking the worst sub-population $\mathbb Q^\star$ from $\mathcal{P}_{\alpha_0}(\mathbb P_{\text{tr}})$, and a \emph{stable learner} learns a more stable representation with smaller discrepancy between $\mathbb Q^\star(Y|\Phi_\theta(X))$ and $\mathbb P_{\text{tr}}(Y|\Phi_\theta(X))$.

 \begin{algorithm}[tb]
   \caption{Stable risk minimization (SRM)}
   \label{algo:DIL}
\begin{algorithmic}
   \STATE {\bfseries Input:} Training Data $D=\{x_i,y_i\}_{i=1}^n$, hyper-parameter $\lambda$, epoch number $T$, prescribed sub-population ratio $\alpha_0$. 
   \STATE {\bfseries Initialize:} $\Phi^{(1)}=X$
   \FOR{ $t=1$ {\bfseries to} $T$}
  	\STATE \textbf{Step 1. Variation explorer}: Given $\Phi^{(t)}$, find the worst sub-population $\mathbb Q^\star (t)$ characterized by ${\bf w}^*$ according to Equation \eqref{equ:empirical_w_star}. 
  	\STATE \textbf{Step 2. Stable learner}: Given the learned worst sub-population $\mathbb Q^\star (t)$, perform stable risk minimization on $\{\hat{\mathbb P}_{\text{tr}}, \mathbb Q^\star (t)\}$ according to Equation \eqref{equ:stable-learner} to obtain the representation $\Phi^{(t+1)}$.
   \ENDFOR
 \end{algorithmic}
  \end{algorithm}

\subsection{Player 1: variation explorer}

Given current representation $\Phi_\theta(X)$ (\emph{abbr.} $\Phi$), the $\alpha_0$-distributional stability takes the form of:
\begin{align*}
	\text{DS}_{\alpha_0}(Y|\Phi;\mathbb P_{\text{tr}}) = \sup\limits_{\mathbb Q\in\mathcal{P}_{\alpha_0}(\mathbb P_{\text{tr}})}\mathbb{E}_{\mathbb Q}\left[\log\frac{\mathbb Q(Y|\Phi)}{\mathbb P_{\text{tr}}(Y|\Phi)}\right].
\end{align*}	
The goal of the variation explorer is to find the sub-population $\mathbb Q^\star$ that:
\begin{equation}
\label{equ:qstar}
	\mathbb Q^\star = \arg\sup\limits_{\mathbb Q\in\mathcal{P}_{\alpha_0}(\mathbb P_{\text{tr}})}\mathbb{E}_{\mathbb Q}\left[\log\frac{\mathbb Q(Y|\Phi)}{\mathbb P_{\text{tr}}(Y|\Phi)}\right].
\end{equation}	
For different kinds of tasks, we propose different ways to approximate Equation \eqref{equ:qstar} in the following.

\emph{\bf (1) For regression tasks}. \quad Given the representation $\Phi\in\Upsilon$ and the label $Y\in\mathbb{R}$, we parameterize the conditional distribution $\mathbb P_{\text{tr}}(Y|\Phi)$ and $\mathbb Q(Y|X)$ as:
\begin{align*}
	& \mathbb P_{\text{tr}}(Y|\Phi) \approx \mathcal{N}(f_{\text{tr}}(\Phi),\sigma_{\text{tr}}^2),\\
	& \mathbb Q(Y|\Phi) \approx \mathcal{N}(f_{\text q}(\Phi),\sigma_{\text q}^2),
\end{align*}
where $f_{\text{tr}}=\mathbb E_{\mathbb P_{\text{tr}}}[Y|\Phi]$ and $f_{\text{q}}=\mathbb E_{\mathbb Q}[Y|\Phi]$ denote the prediction functions tailored to fit the data distributions $\mathbb P_{\text{tr}}$ and $\mathbb Q$, respectively.
$\sigma_{tr},\sigma_q$ are noise scale parameters.
Based on this approximation, Equation \eqref{equ:qstar} for \emph{regression} tasks becomes:
\begin{equation*}
	\mathbb Q^\star = \arg\sup\limits_{\mathbb Q\in\mathcal{P}_{\alpha_0}(\mathbb P_{\text{tr}})}\mathbb{E}_{\mathbb Q}\left[\frac{(Y-f_{\text{tr}}(\Phi))^2}{\sigma_{\text{tr}}^2}-\frac{(Y-f_{\text q}(\Phi))^2}{\sigma_{\text q}^2}\right].
\end{equation*}

\textbf{(2) For classification tasks.}.\quad Denote the number of classes by $K$, the conditional distribution is discrete and can be modeled via a $K$-dimensional simplex.
Given the representation $\Phi\in\Upsilon$ and target variable $Y\in [K]$, $P_{tr}(Y|\Phi)$ and $Q(Y|\Phi)$ are modeled as:
\begin{align*}
	& \mathbb P_{\text{tr}}(Y|\Phi) \approx f_{\text{tr}}(\Phi)\in\Delta_K,\\
	& \mathbb Q(Y|\Phi) \approx f_{\text q}(\Phi)\in\Delta_K,
\end{align*}
where $f_{\text{tr}},f_{\text q}$ denote the prediction models that fit the data from distribution $\mathbb P_{\text{tr}}$ and $\mathbb Q$ respectively.
Then Equation \eqref{equ:qstar} for \emph{classification} tasks becomes:
\begin{equation*}
	\mathbb Q^\star = \arg\sup\limits_{\mathbb Q\in\mathcal{P}_{\alpha_0}(\mathbb P_{\text{tr}})}\mathbb{E}_{\mathbb Q}\left[\log\frac{f_{\text q}(\Phi)[Y]}{f_{\text{tr}}(\Phi)[Y]}\right].
\end{equation*}	
where $f_q(\Phi)[Y]$ denotes the value of $Y$-th dimension of $f_{\text q}(\Phi)\in \Delta_K$, and the same for $f_{\text{tr}}(\Phi)[Y]$.

Now we are ready to derive the empirical objective function from Equation \eqref{equ:qstar} for both regression and classification tasks.
Empirically, given dataset $D=\{x_i,y_i\}_{i=1}^n$ drawn from $\mathbb P_{\text{tr}}$, $\hat{\mathbb P}_{\text{tr}}$ can be represented by 
\begin{equation*}
	\hat{\mathbb P}_{\text{tr}} = \frac{1}{n}\sum_{i=1}^n \delta_{(x_i,y_i)},
\end{equation*}
where $\delta_{(x,y)}$ denotes the Dirac distribution that is supported on $(x,y)$.
Similarly, the sub-population set $\mathcal{P}_{\alpha_0}(\hat{\mathbb P}_{\text{tr}})$ can be modeled as:
\begin{equation*}
	\mathcal{P}_{\alpha_0}(\hat{\mathbb P}_{tr}) = \{{\bf w}=[w_1,\dots,w_n]^T:{\bf w}\in\Delta_n, {\bf w}\leq \frac{1}{\alpha_0n}\},
\end{equation*}
where the sub-population $Q\in \mathcal{P}_{\alpha_0}(\hat{\mathbb P}_{tr})$ is characterized by sample weights, and $w_i$ denotes the weight of the $i$-th sample.
Then Equation \eqref{equ:qstar} can be reformulated as:
\begin{equation}
\label{equ:empirical_w_star}
\begin{aligned}
	&{\bf w}^\star=\arg\max_{{\bf w}\in \mathcal{P}_{\alpha_0}(\hat{\mathbb P}_{tr})}\sum_{i=1}^n w_i \cdot \boxed{g_i}\ \ ,\\
	&\text{s.t. }\quad {\bf w}\in\Delta_n\text{ and }{\bf w}\leq \frac{1}{\alpha_0n},
\end{aligned}
\end{equation}	
where $\boxed{g_i}$ depends on the task type (regression or classification):
\begin{align*}
	\boxed{g_i} := \begin{cases}
		\frac{(y_i-f_{\text{tr}}(\phi_i))^2}{\sigma_{\text{tr}}^2}-\frac{(y_i-f_{\text q}(\phi_i))^2}{\sigma_{\text q}^2},&\ \text{for regression}\\
		\log\frac{f_{\text q}(\phi_i)[y_i]}{f_{\text{tr}}(\phi)[y_i]},&\ \text{for classification}
	\end{cases}
\end{align*}	
where $\phi_i=\Phi(x_i)$.
To estimate $f_{\text{tr}}, \sigma_{\text{tr}}, f_{\text q}, \sigma_{\text q}$, through maximal likelihood estimation, we have:
\begin{align*}
	f_{\text{tr}} &= \arg\min_f \sum_{i=1}^n \ell(f(\phi_i),y_i),\\
	f_{\text q} &= \arg\min_f \sum_{i=1}^n w_i\ell(f(\phi_i),y_i),\\
	\sigma_{\text{tr}}^2 &= \mathbb E_{\mathbb P_{\text{tr}}}[\ell^2(f_{\text{tr}}(\Phi),Y)] - (\mathbb E_{\mathbb P_{\text{tr}}}[\ell(f_{\text{tr}}(\Phi),Y)])^2,\\
	\sigma_{\text q}^2 &= \mathbb E_{\mathbb Q}[\ell^2(f_{\text{q}}(\Phi),Y)] - (\mathbb E_{\mathbb Q}[\ell(f_{\text{q}}(\Phi),Y)])^2.
\end{align*}
Notably, for $f_{\text{tr}}$, since it fits the empirical training distribution $\hat{\mathbb P}_{\text{tr}}$ and is not affected by sample weights, we only train it once and fix it.
For $f_{\text q}$, one could use bi-level optimization to jointly optimize the sample weights $\textbf{w}$ and $f_{\text q}$. 
In this work, we find that adopting an iterative training process yields impressive results in practical, and therefore we did not implement bi-level optimization here.
But we refer the readers interested in the bi-level optimization of Equation \eqref{equ:empirical_w_star} to~\citep{shu2019meta, shaban2019truncated}.

\textbf{Complexity Analysis}\quad Here we analyze the complexity of the variation exploitation stage.
\emph{First}, this stage is based on the \emph{representation} $\Phi(X)$ of the input data $X$. Therefore, the conditional distribution $\mathbb P(Y|\Phi(X))$ is easy to fit empirically and typically is chosen as \emph{linear model}, which could be viewed as the last layer of a deep neural network.
\emph{Second}, we analyze the additional computation cost, and show that it is similar to \emph{adversarial training}.
Denote the sample size as $N$, dimension of $\Phi$ as $d_\phi$, and the training epoch as $T$, the additional cost is $\mathcal O(Nd_\phi T)$. 
Notably, since $f_{\text q}$ is linear, the convergence is quick, and we set it to 50 in our experiments.
To demonstrate that this computation cost is acceptable, we further analyze the additional cost of adversarial training for comparison. Denote the overall number of parameters as $D$, the attack step as $T_a$, the additional cost of adversarial training is $\mathcal O(NDT_a)$ with $D\gg d_\phi$.
Therefore, \emph{the additional computation cost of our method is lower (or no larger) than adversarial training, which is acceptable}.
\emph{Third}, to further lower the computation burden, we perform the variation exploitation stage once every $K$ epochs, and $K$ is set to 20 in our experiments. Therefore, the additional time complexity further reduces to $\mathcal O(Nd_\phi T/K)$.

\subsection{Player 2: stable learner}
\label{sec:step2}

Given the worst sub-population $\mathbb Q^\star$ in Equation \eqref{equ:qstar}, the distributional stability could be simplified to:
\begin{equation*}
	\text{DS}_{\alpha_0}(Y|\Phi;\mathbb P_{\text{tr}}) = \rho_{\text{KL}}(\mathbb Q^\star(Y|\Phi)\|\mathbb P_{\text{tr}}(Y|\Phi)).
\end{equation*}	

Therefore, for the \emph{stable learner} (player 2), the Lagrangian penalty problem in Equation \eqref{equ:lagrangian} becomes:
\begin{equation}
\begin{aligned}
\label{equ:stable-learner}
	\mathcal L(\theta,\eta) =  &\mathbb{E}_{\mathbb P_{\text{tr}}}[\ell(h_\eta(\Phi_\theta(X)),Y)]+\\
	&\lambda\rho_{\text{KL}}(\mathbb Q^\star(Y|\Phi_\theta(X))\|\mathbb P_{\text{tr}}(Y|\Phi_\theta(X))).
\end{aligned}
\end{equation}	
Following the approximation in~\citep{MIP}, we have:
\begin{align*}
	&\rho_{KL}(\mathbb Q^\star(Y|\Phi_\theta(X))\|\mathbb P_{\text{tr}}(Y|\Phi_\theta(X)))\approx \mathcal O(\alpha^2) + \\
	&\quad\quad \alpha \nabla_{\theta,\eta}\left(\mathcal{R}_{\mathbb P_{\text{tr}}}(\theta,\eta)-\mathcal{R}_{\mathbb Q^\star}(\theta,\eta)\right)^T\nabla_{\theta,\eta}\mathcal R_{\mathbb Q^\star}(\theta,\eta)
\end{align*}	
 where $\alpha$ is the learning rate of model parameters $\theta,\eta$, $\mathcal{R}_{\mathbb P_{\text{tr}}}=\mathbb E_{\mathbb P_{\text{tr}}}[\ell(X,Y)]$ denotes the average prediction error under distribution $\mathbb P_{\text{tr}}$, and $\mathcal{R}_{\mathbb Q^\star}=\mathbb E_{\mathbb Q^\star}[\ell(X,Y)]$ denotes the average prediction error under distribution $\mathbb Q^\star$. 
 
Given the worst sub-population $\mathbb Q^\star$, the overall objective function of the player 2 becomes:
\begin{equation}
\begin{aligned}
\label{equ:stable-learner}
	\mathcal L&(\theta,\eta) =  \mathbb{E}_{\mathbb P_{\text{tr}}}[\ell(h_\eta(\Phi_\theta(X)),Y)]+\\
	&\lambda\nabla_{\theta,\eta}\left(\mathcal{R}_{\mathbb P_{\text{tr}}}(\theta,\eta)-\mathcal{R}_{\mathbb Q^\star}(\theta,\eta)\right)^T\nabla_{\theta,\eta}\mathcal R_{\mathbb Q^\star}(\theta,\eta),
\end{aligned}
\end{equation}	
which can be efficiently optimized via gradient descent.

\section{RELATED WORK}
In this section, we discuss the related works in detail.
There are mainly two branches of literatures related to our work, including invariant learning~\citep{IRM,IRMgames,MIP,hrm,KerHRM,IB,EIIL} and distributionally robust optimization~\citep{fDRO, Certifying}.

For invariant learning, \citet{IRM} first come up with the OOD generalization problem and design a regularizer to learn such representations that the optimal linear classifier remains the same across training environments, and this method is a typical method in invariant learning.
And \citet{MIP} theoretically characterize when the invariance will benefit OOD generalization and propose to learn the maximal invariant predictor to achieve OOD optimality.
\citet{IB} combines invariant learning with information bottleneck for better OOD generalization performance.
The proposed invariance definition requires an invariant relationship among all possible environments, termed as the strict invariance.
However, whether it exists in real applications remains doubtful, since the noises are likely to change in different environments and therefore violate the strict invariance.
Further, the availability of multiple training environments itself is quite hard to meet with in real scenarios, making many invariant learning methods inapplicable in real applications.

In order to mitigate such limitations, recently, some works~\citep{EIIL,hrm,KerHRM} try to learn pseudo-environments first and then perform invariant learning.
\citet{EIIL} directly maximize the regularizer of IRM with a given biased model to generate environments.
\citet{hrm, KerHRM} propose to iteratively learn the environment splits and the invariant predictors, although intuitively reasonable, the property of learned environments still remains vague, which renders the proposed framework unstable.
Since the property of learned environments cannot be analyzed or guaranteed, whether the invariance can be achieved also remains unclear and cannot be certified.
Inspired by this, it is of paramount importance to reformulate the invariant learning problem under latent heterogeneity to a more reasonable one.

Distributionally robust optimization (DRO) methods, typified by $f$-DRO~\citep{fDRO}, propose to optimize the worst-case error with respect to a pre-defined distribution set that lies around the training distribution.
When the testing distribution lies in the pre-defined distribution set, the OOD generalization performance can be controlled by the worst-case.
However, when the target distribution is not captured by the pre-defined set, the performance of DRO depends on the relationship between the target distribution and the worst-case distribution in the pre-defined set, which cannot be guaranteed. 
This is also reflected in our Figure \ref{fig:certified} (the curve of $f$-DRO is quite fluctuant).
Unfortunately, such circumstances are quite likely to happen in real scenarios, since the pre-defined set cannot be set too large because of the over-pessimism problem~\citep{Hudoes,UnlabelDRO}.
In this work, we borrow the idea of distribution set from DRO to characterize the sub-population set, based on which we come up with the notion of distributional stability, which is a relaxed alternative of the strict invariance.

\begin{table*}[htbp]
	\centering
	\vspace{-1em}
	\caption{Overall results in selection bias simulation experiments with varying bias rates $r_1$.}
	\label{tab:sim_selection}
	
	\resizebox{0.95\textwidth}{15mm}{
	\begin{tabular}{lccccccccc}
		\toprule
		Bias Ratio $r$&\multicolumn{3}{c}{$r_1=1.5$}&\multicolumn{3}{c}{$r_1=1.9$}&\multicolumn{3}{c}{$r_1=2.3$}\\
		\midrule
		Methods &  $\mathrm{Mean\ Error}$ & $\mathrm{Std\ Error}$& $\mathrm{Max\ Error}$&$\mathrm{Mean\ Error}$ & $\mathrm{Std\ Error}$& $\mathrm{Max\ Error}$ &  $\mathrm{Mean\ Error}$ & $\mathrm{Std\ Error}$& $\mathrm{Max\ Error}$ \\
		ERM &2.651\scriptsize($\pm 0.106$)  &0.119\scriptsize($\pm 0.038$)  &2.820\scriptsize($\pm 0.140$) &3.155\scriptsize($\pm 0.210$) &0.147\scriptsize($\pm 0.039$) &3.348\scriptsize($\pm 0.184$) &3.240\scriptsize($\pm 0.174$) &0.136\scriptsize($\pm 0.039$) &3.433\scriptsize($\pm 0.197$) \\
		$f$-DRO&1.835\scriptsize($\pm 0.144$) &0.070\scriptsize($\pm 0.024$) &1.940\scriptsize($\pm 0.169$)  &1.973\scriptsize($\pm 0.261$)  &0.096\scriptsize($\pm 0.025$) &2.107\scriptsize($\pm 0.274$) &2.018\scriptsize($\pm 0.422$) &0.100\scriptsize($\pm 0.025$) &2.149\scriptsize($\pm 0.425$)  \\
		EIIL &1.764\scriptsize($\pm 0.402$) &0.074\scriptsize($\pm 0.022$) &1.864\scriptsize($\pm 0.423$)  &2.043\scriptsize($\pm 0.600$)  &0.101\scriptsize($\pm 0.036$) &2.185\scriptsize($\pm 0.656$) &1.840\scriptsize($\pm 0.347$) &0.085\scriptsize($\pm 0.022$) &1.962\scriptsize($\pm 0.349$) \\
		KerHRM &1.825\scriptsize($\pm 0.354$)  &0.089\scriptsize($\pm 0.040$) &1.978\scriptsize($\pm 0.374$)  &1.658\scriptsize($\pm 0.472$) &0.068\scriptsize($\pm 0.031$) &1.788\scriptsize($\pm 0.617$) &1.572\scriptsize($\pm 0.504$) &0.088\scriptsize($\pm 0.036$) &1.677\scriptsize($\pm 0.537$)\\
		IRM(with $\mathcal{E}_{tr}$ label) &1.683\scriptsize($\pm 0.201$) &0.066\scriptsize($\pm 0.024$)&1.780\scriptsize($\pm 0.227$)  &1.782\scriptsize($\pm 0.134$)  &0.067\scriptsize($\pm 0.018$) &1.886\scriptsize($\pm 0.163$)  &1.964\scriptsize($\pm 0.276$) &0.067\scriptsize($\pm 0.015$) &2.057\scriptsize($\pm 0.295$)\\
		SRM   &\bf 1.288\scriptsize($\pm 0.344$)  &\bf 0.059\scriptsize($\pm 0.024$)  &\bf 1.367\scriptsize($\pm 0.365$) &\bf 1.323\scriptsize($\pm 0.223$) &\bf 0.054\scriptsize($\pm 0.020$) &\bf 1.402\scriptsize($\pm 0.233$) &\bf 1.382\scriptsize($\pm 0.283$) &\bf 0.059\scriptsize($\pm 0.018$) &\bf 1.457\scriptsize($\pm 0.299$) \\
		\bottomrule
	\end{tabular}
	}
	\vspace{-1em}
\end{table*}

\begin{table*}[htbp]
	\centering
	\caption{Overall results in classification simulation experiments with varying bias rates $r_2$.}
	\label{tab:sim1}
	
	\resizebox{0.95\textwidth}{17mm}{
	\begin{tabular}{lcccccccc}
		\toprule
		Bias Ratio $r_2$&\multicolumn{4}{c}{$r_2=0.75$}&\multicolumn{4}{c}{$r_2=0.80$}\\
		\midrule
		Dimension $d$&\multicolumn{2}{c}{$d=5$} & \multicolumn{2}{c}{$d=10$}&\multicolumn{2}{c}{$d=5$} & \multicolumn{2}{c}{$d=10$}\\
		\midrule
		Methods &  $\mathrm{Train\ Acc}$ & $\mathrm{Test\ Acc}$&  $\mathrm{Train\ Acc}$ & $\mathrm{Test\ Acc}$&  $\mathrm{Train\ Acc}$ & $\mathrm{Test\ Acc}$&  $\mathrm{Train\ Acc}$ & $\mathrm{Test\ Acc}$ \\
		ERM &\bf 0.917\scriptsize($\pm 0.009$) &0.388\scriptsize($\pm 0.039$) &\bf 0.972\scriptsize($\pm 0.007$) &0.573\scriptsize($\pm 0.026$) &\bf 0.931\scriptsize($\pm 0.005$) &0.364\scriptsize($\pm 0.023$) &\bf 0.975\scriptsize($\pm 0.005$) &0.526\scriptsize($\pm 0.030$)    \\
		$f$-DRO&0.766\scriptsize($\pm 0.012$) &0.452\scriptsize($\pm 0.021$) &0.920\scriptsize($\pm 0.006$) &0.611\scriptsize($\pm 0.028$) &0.787\scriptsize($\pm 0.011$) &0.427\scriptsize($\pm 0.022$) &0.930\scriptsize($\pm 0.005$) &0.616\scriptsize($\pm 0.022$)    \\
		EIIL &0.727\scriptsize($\pm 0.145$) &0.544\scriptsize($\pm 0.058$) &0.814\scriptsize($\pm 0.160$) &0.451\scriptsize($\pm 0.049$) &0.743\scriptsize($\pm 0.155$) &0.571\scriptsize($\pm 0.050$) &0.823\scriptsize($\pm 0.165$) &0.406\scriptsize($\pm 0.056$)    \\
		KerHRM &0.784\scriptsize($\pm 0.035$) &0.636\scriptsize($\pm 0.182$) &0.834\scriptsize($\pm 0.143$) &0.659\scriptsize($\pm 0.205$) &0.780\scriptsize($\pm 0.043$) &0.665\scriptsize($\pm 0.178$) &0.800\scriptsize($\pm 0.097$) &0.674\scriptsize($\pm 0.139$)    \\
		IRM(with $\mathcal{E}_{tr}$ label) &0.855\scriptsize($\pm 0.010$) &0.467\scriptsize($\pm 0.046$) &0.908\scriptsize($\pm 0.007$) &0.529\scriptsize($\pm 0.058$) &0.876\scriptsize($\pm 0.005$) &0.386\scriptsize($\pm 0.047$) &0.914\scriptsize($\pm 0.006$) &0.448\scriptsize($\pm 0.056$)    \\
		SRM &0.781\scriptsize($\pm 0.032$) &\bf 0.716\scriptsize($\pm 0.066$) &0.869\scriptsize($\pm 0.023$) &\bf 0.684\scriptsize($\pm 0.052$) &0.787\scriptsize($\pm 0.030$) &\bf 0.703\scriptsize($\pm 0.073$) &0.871\scriptsize($\pm 0.017$) &\bf 0.697\scriptsize($\pm 0.061$)    \\
		\bottomrule
	\end{tabular}
	}
\end{table*}

\section{EXPERIMENTS}
\textbf{Baselines}.\\ 
We compare our proposed SRM algorithm with the following methods: Empirical Risk Minimization (ERM), Distributionally Robust Optimization ($f$-DRO,~\citet{fDRO}), Environment Inference for Invariant Learning (EIIL,~\citet{EIIL}), Kernelized Heterogeneous Risk Minimization (KerHRM,~\citet{KerHRM}) and Invariant Risk Minimization (IRM,~\citet{IRM}) with environment $\mathcal{E}_{tr}$ labels.
Note that IRM requires environment labels, and we provide the ground-truth sub-population labels for IRM.

\textbf{Evaluation Metrics}.\\ 
For experiments with multiple testing distributions, we use $\mathrm{Mean}\ \text{Error}$ defined as:
\begin{equation*}
	\mathrm{Mean}\ \text{Error}=\frac{1}{|\mathcal{E}_{\text{test}}|}\sum_{e\in\mathcal{E}_{\text{test}}}\mathbb E_{\mathbb P^e}[\ell(X,Y)],
\end{equation*}
and $\mathrm{Std}\ \text{Error}$ defined as: 
\begin{small}
\begin{equation*}
	\mathrm{Std}\ \text{Error}=\sqrt{\frac{1}{|\mathcal{E}_{\text{test}}|-1}\sum_{e\in\mathcal{E}_{\text{test}}}(\mathbb E_{\mathbb P^e}[\ell(X,Y)]-\mathrm{Mean}\ \text{Error})^2},
\end{equation*}
\end{small}
\noindent and $\mathrm{Max}\ \text{Error}$ defined as:
\begin{equation*}
	\mathrm{Max}\ \text{Error}=\max_{e\in\mathcal{E}_{\text{test}}}\mathbb E_{\mathbb P^e}[\ell(X,Y)],
\end{equation*}
which are mean error, standard deviation error, and the worst-case error across testing environments $\mathcal{E}_{\text{test}}$.

\subsection{Simulation Data}
\label{sec:simulation}

\textbf{Regression with Selection Bias}\\
In this setting, the relationships between covariates and the target are perturbed through the selection bias mechanism across sub-populations.
We generate the data following the mechanism adopted by~\citet{KerHRM,liu2022measure}, where we assume $X=[S,V]^T\in \mathbb R^{10}$ and $\small Y = f(S) + \epsilon = \beta^TS + S_1S_2S_3+\mathcal{N}(0,0.1)$. 
To generate different sub-populations, we maintain $\mathbb P(Y|S)$ the same across sub-populations and leverage a data selection mechanism to vary $\mathbb P(Y|V)$.
Specifically, we select data point $(x_i,y_i)$ with probability $\tau_i$ according to one certain variable $V_b \in V$ as $\tau_i = |r|^{-5 * |y_i - \text{sign}(r)\cdot V_b|}$ where $|r|>1$.
Intuitively, $r$ controls the strengths and direction of the spurious correlation between $V_b$ and $Y$.
The larger value of $|r|$ means the stronger spurious correlation between $V_b$ and $Y$ , and $r>0$ means positive correlation and vice versa (i.e. if $r > 0$, a data point whose $V_b$ is close to its $y$ is more probably to be selected.).
Therefore, we use $r$ to define different sub-populations.

For training data, we mix 2000 data points from different $r_1$ and 200 points from $r_2=-1.1$.
For different testing scenarios, we sample 1000 data points from $r\in\{-1.9, -2.1, \dots, -2.9\}$, respectively.
For our SRM algorithm and $f$-DRO, we set $\alpha_0=0.1$ (the ground truth is 0.09).
Linear models are used in this experiment.

\begin{figure*}[t]
       \subfigure[Certified robustness.]{\label{fig:certified}\includegraphics[width=0.33\textwidth,height=1.5in]{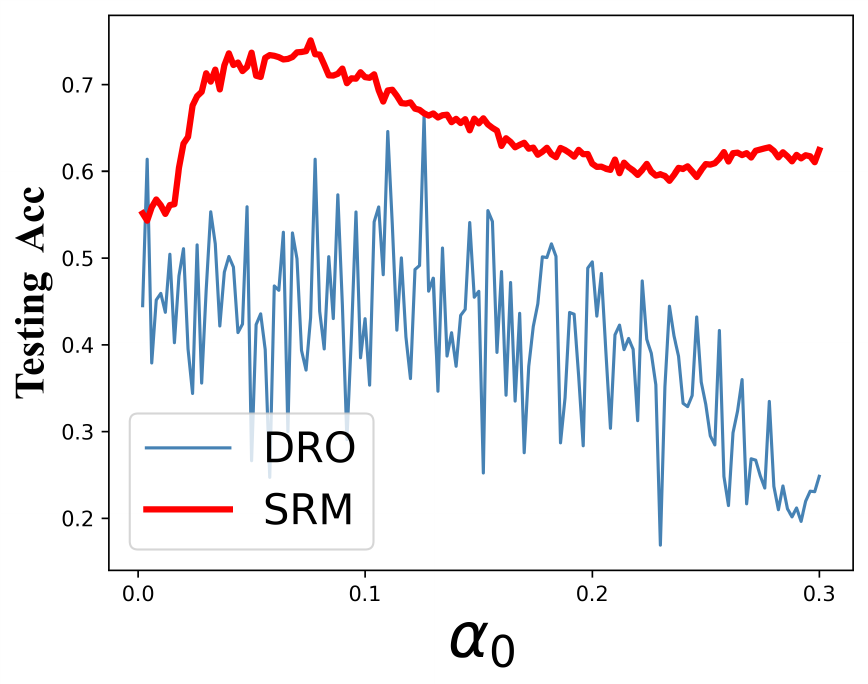}}
       \subfigure[F1 Score and Testing Accuracy.]{\label{img:scatter}\includegraphics[width=0.33\textwidth,height=1.5in]{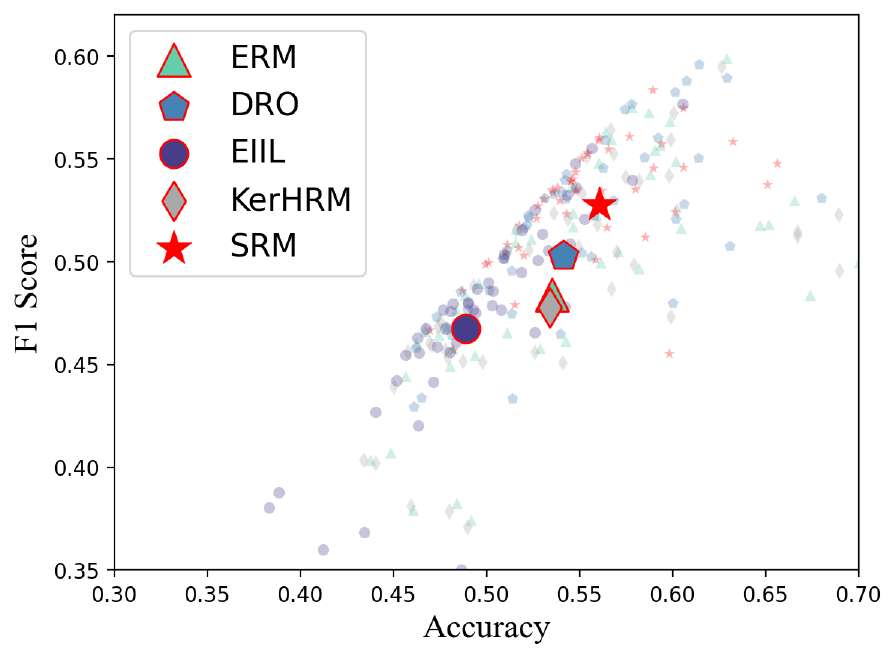}}
		\subfigure[Overall testing accuracy.]{\label{img:distribution}\includegraphics[width=0.33\textwidth,height=1.5in]{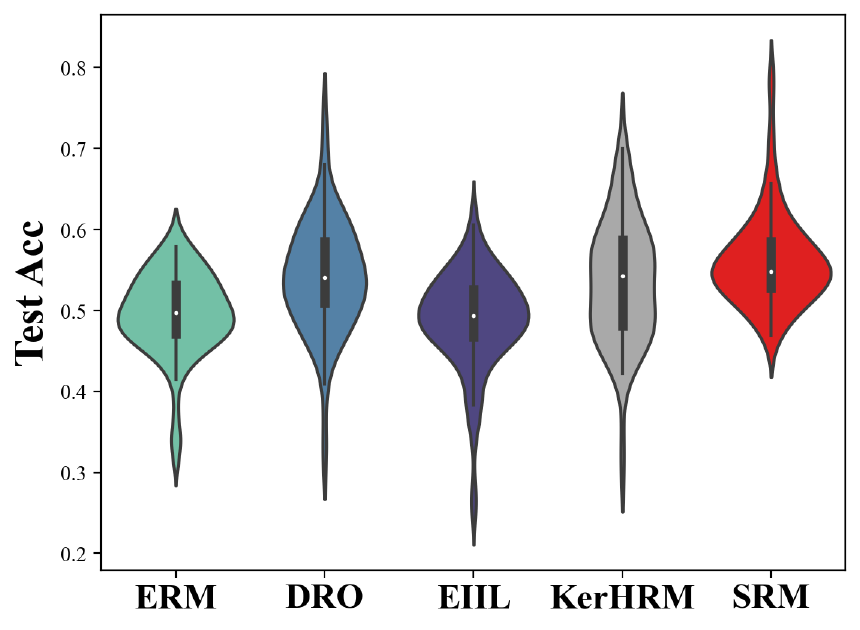}}       
		\caption{Experimental results. (a): Demonstration of the certified robustness via the classification task (in Section \ref{sec:simulation}), where we vary the $\alpha_0$ and plot the corresponding testing accuracy for $f$-DRO and our proposed SRM. (b): The F1 score and testing accuracy on all 50 target states of different methods. We highlight the average F1 score and testing accuracy (in Section \ref{sec:realdata}). (c): The distribution of testing accuracy of different methods (in Section \ref{sec:realdata}).}
		\label{fig:bar}
\end{figure*}

\textbf{Classification with Spurious Correlation}\\
Following \cite{Classification}, we induce spurious correlations between the label $Y\in\{+1,-1\}$ and a spurious attribute $A\in\{+1,-1\}$ of different strengths and directions.
We assume $X=[S,V]^T\in\mathbb{R}^{2d}$, where $S\in\mathbb{R}^d$ is the invariant feature generated from the label $Y$ and $V \in \mathbb{R}^d$ the variant feature generated from the spurious attribute $A$:
\begin{equation}
	S|Y\sim\mathcal{N}(Y\textbf{1}, \sigma_s^2\mathbb{I}_d),V|A\sim\mathcal{N}(A\textbf{1}, \sigma_v^2\mathbb{I}_d). 
\end{equation}
In this setting, we characterize different groups with the bias rate $r\in(0,1]$, which represents that for $100\cdot r\%$ data, $A=Y$, and for the other $100\cdot(1-r)\%$ data, $A=-Y$. 
Intuitively, $r$ controls the spurious correlation between the label $Y$ and spurious attribute $A$.
In training, we generate 2000 data points, where $50\%$ points are from group 1 with $r_1=0.9$ and the other from group 2 with varying $r_2$.
In testing, we generate 1000 data points with $r_3=0.0$ to simulate strong distributional shifts, since the direction of spurious correlations is reversed from training. 
We design multiple settings with different bias rates $r_2$ as well as the dimensions $d$ of features.
For our SRM algorithm and $f$-DRO, we set $\alpha_0=0.15$ (the ground truth is 0.17).
We use a two-layer MLP for this experiment.

\textbf{Better OOD Generalization Performance: }We report the results of the regression and classification tasks in Table \ref{tab:sim_selection} and \ref{tab:sim1}.
From the results, our SRM outperforms all baselines in terms of higher prediction accuracy and better stability among distributional shifts, which validates that our SRM can achieve better OOD generalization performance and is consistent with our theoretical analysis in Theorem \ref{theorem:bound}.

\textbf{$\alpha_0$ Controls the Extent of Stability:}
In the definition of $\alpha_0$-distributional stability, $\alpha_0$ controls the range of stability, i.e. smaller $\alpha_0$ examines more fine-grained stability.
To demonstrate the effect of $\alpha_0$ in our SRM algorithm, for the classification task, we plot the curve of testing accuracy w.r.t. $\alpha_0$ for our SRM and $f$-DRO in Figure \ref{fig:certified}. 
Since the real proportion of the minor sub-population is set to 0.17, we hope SRM is effective when $\alpha_0\leq 0.17$.
From the results, we could see that the performances of SRM maintain at a high level for $\alpha_0\in [0.05,0.17]$, which validates our intuitions.
For too small $\alpha_0$, the performances drop due to the insufficient number of samples and stronger noises.
Also, the performances of $f$-DRO are oscillating, which corresponds with our analysis in Remark \ref{remark} that: since distributional robustness only cares about the worst sub-population performances, when the testing distribution falls out of the pre-defined distribution set, it cannot guarantee the OOD generalization performance.
However, for our SRM, the guarantees for the OOD generalization ability in Theorem \ref{theorem:bound} do not put strong requirements for the testing distributions, since it only requires the learnability of the problem.

\subsection{Real-World Data: Retiring Adults}
\label{sec:realdata}
To better validate the effectiveness of the proposed SRM algorithm, we consider a much more challenging scenario on a real-world dataset, named ACSTravelTime~\citep{OODfair}.
The task is to predict whether an individual has a commute to work that is longer than 20 minutes.
In this task, we have 16 features and 1,428,642 data points in total from all 50 US states.
Since there are 50 distinct environments, this dataset contains natural geographic shifts, which makes it suitable for testing the OOD generalization performances.
In \emph{training}, we sample 2000 data points from MA and validate on the rest data from MA.
In \emph{testing}, we test different methods on all the other 49 states.

In Figure \ref{img:scatter}, we plot the accuracy and F1 score for each method on the 50 states, and in Figure \ref{img:distribution} we show the overall testing accuracy of different methods.
Note that the original code released by KerHRM is too time-consuming to run on this data because of the large amount of data (over 1 million data points), therefore we use HRM~\citep{hrm} here to replace the KerHRM, which can only deal with the raw feature data.
Since there is one environment in this experiment and we do not know the underlying sub-populations, we cannot compare with IRM in this setting.
And EIIL can be viewed as an alternative to IRM with learned environments from training data.

From the results in Figure \ref{img:scatter}, the average performance of our SRM locates in the top right of the figure, which shows that our methods achieve the best OOD generalization performance w.r.t. testing accuracy and F1 score.
Further, in Figure \ref{img:distribution}, for our SRM, the performances of most environments are concentrated at high accuracy, and the variance of different environments is significantly smaller than the baselines.
It shows that our SRM algorithm can learn some distributional stability among different sub-populations, which benefits the generalization performances.
And the good OOD generalization performance also corresponds with our intuition from Theorem \ref{theorem:bound} that considering the distributional stability could benefit the OOD generalization error.

\section{CONCLUSION}
In this paper, we propose the distributional stability, which measures the stability of prediction mechanisms among sub-populations.
Based on this criterion, we propose an approximated algorithm, termed stable risk minimization, to enhance the model's stability with respect to distribution shifts in prediction mechanisms.
Despite the theoretical and empirical results, our work has the following limitations (or potential directions to improve):

\textbf{Analysis of the approximation.}\quad 
Based on the overall objective function in Equation \eqref{equ:objective}, we make several approximations to derive a tractable optimization algorithm. 
A notable challenge associated with this approach is the difficulty in thoroughly analyzing the behavior of the approximated algorithm, particularly with regard to its convergence properties and the bounds on its generalization error.
A promising avenue for future research lies in the development of improved approximation techniques that come with stronger theoretical guarantees.

\textbf{Lack of large-scale suitable datasets.}\quad 
In the current version of our study, both simulated and real-world experiments are conducted on a small scale. This limitation is largely due to the nature of datasets commonly employed in large-scale research, which predominantly consist of image data. 
These datasets usually exhibit shifts in the input space, $X$, rather than in the conditional distribution ($Y|X$-shifts) that are more pertinent to our investigation into invariant learning.

As the field of invariant learning evolves, a noticeable trend is the application of these methods to complex tasks, particularly image classification datasets. 
However, a crucial question emerges: Are these image datasets genuinely conducive to invariant learning methods aimed at aligning the $Y|X$ distributions? 
Research by~\cite{gulrajani2020search} reveals that Empirical Risk Minimization (ERM) often outperforms most domain generalization and invariant learning methods tailored for these datasets. 
This suggests that the prevalent distribution shifts in image datasets are primarily $X$-shifts, with the primary objective being to model $\mathbb{E}_{\mathbb{P}_{\text{tr}}}[Y|X]$.
Additionally, numerous empirical studies, such as those by~\citep{miller2021accuracy}, have identified a strong correlation between out-of-distribution (OOD) generalization performance and in-distribution (ID) performance. 
This correlation further underscores the inadequacy of traditional image classification tasks as a testing ground for invariant learning methods.

In light of these findings, we advocate for a shift in research focus towards understanding the patterns of distribution shifts in real-world applications, as highlighted by~\citep{liu2023need}. 
A promising avenue of exploration involves the creation of real-world, large-scale datasets featuring $Y|X$-shifts. 
These datasets would likely offer a more fitting and challenging environment for assessing the capabilities of invariant learning methods.

\newpage

\section{Acknowledgements}
Peng Cui was supported by National Natural Science Foundation of China (No. 62141607).
Bo Li's research was supported by the National Natural Science Foundation of China (No.72171131, 72133002); the Technology and Innovation Major Project of the Ministry of Science and Technology of China under Grants 2020AAA0108400 and 2020AAA0108403.

\bibliography{example_paper}
\bibliographystyle{apalike}

\section*{Checklist}

 \begin{enumerate}

 \item For all models and algorithms presented, check if you include:
 \begin{enumerate}
   \item A clear description of the mathematical setting, assumptions, algorithm, and/or model. [Yes]
   \item An analysis of the properties and complexity (time, space, sample size) of any algorithm. [Not Applicable]
   \item (Optional) Anonymized source code, with specification of all dependencies, including external libraries. [Not Applicable]
 \end{enumerate}

 \item For any theoretical claim, check if you include:
 \begin{enumerate}
   \item Statements of the full set of assumptions of all theoretical results. [Yes]
   \item Complete proofs of all theoretical results. [Yes]
   \item Clear explanations of any assumptions. [Yes]     
 \end{enumerate}

 \item For all figures and tables that present empirical results, check if you include:
 \begin{enumerate}
   \item The code, data, and instructions needed to reproduce the main experimental results (either in the supplemental material or as a URL). [Yes]
   \item All the training details (e.g., data splits, hyperparameters, how they were chosen). [Yes]
         \item A clear definition of the specific measure or statistics and error bars (e.g., with respect to the random seed after running experiments multiple times). [Yes]
         \item A description of the computing infrastructure used. (e.g., type of GPUs, internal cluster, or cloud provider). [Yes]
 \end{enumerate}

 \item If you are using existing assets (e.g., code, data, models) or curating/releasing new assets, check if you include:
 \begin{enumerate}
   \item Citations of the creator If your work uses existing assets. [Yes]
   \item The license information of the assets, if applicable. [Not Applicable]
   \item New assets either in the supplemental material or as a URL, if applicable. [Not Applicable]
   \item Information about consent from data providers/curators. [Not Applicable]
   \item Discussion of sensible content if applicable, e.g., personally identifiable information or offensive content. [Not Applicable]
 \end{enumerate}

 \item If you used crowdsourcing or conducted research with human subjects, check if you include:
 \begin{enumerate}
   \item The full text of instructions given to participants and screenshots. [Not Applicable]
   \item Descriptions of potential participant risks, with links to Institutional Review Board (IRB) approvals if applicable. [Not Applicable]
   \item The estimated hourly wage paid to participants and the total amount spent on participant compensation. [Not Applicable]
 \end{enumerate}

 \end{enumerate}

\newpage
\appendix

\section{PROOF}

\label{appendix:theorem}
\begin{proof}
	Denote the upper bound of $\ell(\cdot,\cdot)$ as $M>0$. 
	For any $e\in\text{supp}(\mathcal{E})$,	denote $\mathbb P^{'}_{e}(Y,\Phi)=\mathbb P^e(Y|\Phi)\mathbb P(\Phi)$ and $\mathbb P^{'}_{\text{tr}}(Y,\Phi)=\mathbb P_{\text{tr}}(Y|\Phi)\mathbb P(\Phi)$, and then we have
	\begin{align}
		&\mathbb{E}\left[\mathbb{E}_{\mathbb P^e}[\ell(f(\Phi),Y)|\Phi]-\mathbb{E}_{\mathbb P_{\text{tr}}}[\ell(f(\Phi),Y)|\Phi]\right]\\
		\leq\quad& 2M\cdot \text{TV}(\mathbb P^{'}_{e}, \mathbb P^{'}_{\text{tr}})\\
		\leq\quad& 2M\cdot \sqrt{\frac{1}{2}\rho_{\text{KL}}(\mathbb P^{e}(Y|\Phi)\|\mathbb P^{\text{tr}}(Y|\Phi))}\\
		\leq\quad & \mathcal O(\sqrt{1-e^{-s(\delta)}})
	\end{align}
\end{proof}

\section{EXPERIMENTAL DETAILS}
\label{appendix:exp}
In this section, we demonstrate the details of our simulated experiments.

\textbf{Regression}\quad 
In this setting, the correlations among covariates are perturbed through a selection bias mechanism. 
We assume $X = [S,V]^T \in \mathbb{R}^{10}$ with $S \in \mathbb{R}^{5}$ and $V \in \mathbb{R}^{5}$. 
We assume $Y = f(S) + \epsilon$ and $\mathbb P(Y|S)$ remains invariant across environments while $P(Y|V)$ can arbitrarily change. 

Therefore, we generate training data points with the help of auxiliary variables $Z \in \mathbb{R}^{6}$ as following:
\begin{align}
&Z_1, \dots, Z_{6} \stackrel{iid}{\sim} \mathcal{N}(0,2.0) \\
&V_1, \dots, V_{5} \stackrel{iid}{\sim} \mathcal{N}(0,2.0) \\
&S_i = 0.8*Z_i + 0.2 * Z_{i+1} \ \ \ \ \ for \ \ i = 1, \dots, 5
\end{align}
To induce model misspecification, we generate $Y$ as:
\begin{equation}
Y = f(S) + \epsilon = \theta_s(S)^T + S_1S_2S_3+\epsilon
\end{equation}
where $\theta_s = [\frac{1}{2},-1, 1, -\frac{1}{2}, 1]$, and $\epsilon \sim \mathcal{N}(0, 1.0)$. 
As we assume that $\mathbb P(Y|S)$ remains unchanged while $\mathbb P(Y|V)$ can vary across environments, we design a data selection mechanism to induce this kind of distribution shifts.
For simplicity, we select data points according to a certain variable $V_b \in V$:
\begin{align}
&\tau =|r|^{-5*|y - \text{sign}(r)*v_b|}  \\
&\mu \sim Uni(0,1 ) \\
&M(r;(x,y)) =
\begin{cases}
1, \ \ \ \ \ &\text{$\mu \leq \tau$ } \\
0, \ \ \ \ \ &\text{otherwise}
\end{cases} 
\end{align}  
where $|r| > 1$.
Given a certain $r$, a data point $(x,y)$ is selected if and only if $M(r;(x,y))=1$ (i.e. if $r>0$, a data point whose $V_b$ is close to its $Y$ is more probably to be selected.)

\textbf{Classification}\quad We set $\sigma_s^2=3.0$ and $\sigma_v^2=0.3$ to let the model more prone to use spurious $V$ since it is more informative.

As for the hyper-parameter for SRM and $f$-DRO, for regression data, we set $\alpha_0=0.1$ (the true minor subpopulation ratio is 0.09); for classification data, we set $\alpha_0=0.15$ (the true minor subpopulation ratio is 0.17).
As for the validation data, we sample $i.i.d$ data as training data and compare both the worst-case performance of two subpopulations.
As for IRM, we select the parameter of the regularizer $\lambda \in \{0.1, 0.3, \dots, 0.9, 1.5, 5.0, 10.0\}$ according to the validation performance.
As for EIIL, we set the epochs for splitting environments to 1e4 for good convergence, and other parameters are the same as IRM.
As for KerHRM, we set the cluster\_num to be the ground-truth 2.
All experiments are run on a GPU server with one NVIDIA GeForce RTX 3090.

\end{document}